# THE RECOVERY OF CAUSAL POLY-TREES FROM STATISTICAL DATA *

**George Rebane & Judea Pearl**
Cognitive Systems Laboratory
Computer Science Department
University of California, Los-Angeles, CA. 90024

## ABSTRACT

Poly-trees are singly connected causal networks in which variables may arise from multiple causes. This paper develops a method of recovering poly-trees from empirically measured probability distributions of pairs of variables. The method guarantees that, if the measured distributions are generated by a causal process structured as a poly-tree then the topological structure of such tree can be recovered precisely and, in addition, the causal directionality of the branches can be determined up to the maximum extent possible. The method also pinpoints the minimum (if any) external semantics required to determine the causal relationships among the variables considered.

## 1. INTRODUCTION

The importance of recovering dependency structures from empirical observations is well established in many areas of machine learning. In recent years, the theory of Bayesian Networks has provided powerful techniques (Pearl [1],[2]) for representing and operating on knowledge in the form of directed acyclic graphs (DAGs), describing causal influences among uncertain events.

In this class of dependency models, a poly-tree is a singly connected causal net that includes variables with multiple causes. This structure can be used to model numerous real-world processes and permits evidential reasoning to be conducted coherently and efficiently by concurrent local computations. This paper develops a method to recover poly-trees from empirical observations in the form of a discrete joint probability density function (JPDF).

The method extends the work of Chow and Liu [3] who showed how to recover an undirected Markov tree from a given discrete JPDF using the maximum weight spanning tree (*MWST*) algorithm. Specifically, we prove that the same algorithm also recovers the undirected skeleton (topology) of a poly-tree that would faithfully represent the measured JPDF (if such a representation exists). We then develop an algorithm to recover the causal directionality of the branches from the JPDF to the maximum extent allowed by probability theory.

## 2. THE CHOW & LIU *MWST* ALGORITHM

### 2.1 The Problem Statement

Chow and Liu were interested in estimating an underlying N-dimensional JPDF, $P(X)$, from finite observations and representing it in a parsimonuous way. They developed a computationally efficient means of deriving an approximating distribution $P_a(X)$ of tree-like dependence, thereby limiting $P_a$ to be a product of second-order distributions. Specifically

---
* This work was supported in part by the National Science Foundation Grant, DCR 83-13875.

$$P_a(X) = \prod_i P(x_i \mid x_{j(i)}) \tag{1}$$

where $X = (x_1 x_2, \ldots, x_N)$, $x_1$ is the root of the tree and $x_{j(i)}$ is the parent of $x_i$. The measure of closeness between $P_a$ and $P$ was selected as the average mutual information between the two distributions

$$I(P, P_a) = \sum_X P(X) \log \frac{P(X)}{P_a(X)} \tag{2}$$

first introduced by Lewis[4] for product forms of $P_a$.

## 2.2 Summary of the *MWST* Algorithm

Using the average mutual information between two variables $x_i$ and $x_j$

$$I(x_i, x_j) = \sum_{x_i, x_j} P(x_i, x_j) \log \frac{P(x_i, x_j)}{P(x_i) P(x_j)} \geq 0, \tag{3}$$

as the weight on the branch $(x_i, x_j)$, the minimum value of the proximity measure (2) was shown to be achieved by a distribution $P_a$ that corresponds to a maximum weight spanning tree (*MWST*).

The Chow and Liu *MWST* algorithm is then summarized in the following steps:

1.  From the given (observed) $P(X)$ compute all possible $N(N-1)/2$ branch weights and order them by magnitude.

2.  Assign the largest two branches to the tree to be constructed.

3.  Examine the next largest branch and add to the tree if it does not form a loop, else discard and examine the next largest branch.

4.  Repeat step three until $N - 1$ branches have been selected (a spanning tree constructed.)

5.  $P_a(X)$ can be computed by selecting an arbitrary root node and forming the product in (1).

## 2.3 The Virtues of the *MWST* Algorithm

The virtues of this algorithm are that it only uses second order statistics which are easily and reliably measured and are economical to store. The tree is developed with $O(N^2)$ effort using only numerical comparisons, thereby avoiding expensive tests for conditional independence. It is further shown that if the branch weights are computed from sampled data, then $P_a$ will be a maximum likelihood estimate of $P$. The consequence of this is that if the underlying distribution is indeed one of tree dependence, then the approximating tree recovered by the *MWST* algorithm converges with probability one to the true dependence tree, while the expansion in (4) is applicable to any $P$, the unique structure of poly-trees.

## 3. THE RECOVERY OF POLY-TREES

### 3.1 The Problem Statement

Assume we are given a distribution $P(X)$ of $N$ discrete-valued variables which represents the JPDF of a specific but unknown *generating poly-tree* (GPT). In other words, $P(X)$ is given by

$$P(X) = \prod_{i=1}^{N} P(x_i \mid x_{j_1(i)}, x_{j_2(i)}, \cdots, x_{j_m(i)}) \tag{4}$$

where $\{x_{j_1(i)}, x_{j_2(i)}, \cdots, x_{j_m(i)}\}$ is the (possibly empty) set of direct parents of variable $x_i$ in the GPT and, moreover, the parents of each variable are mutually independent, i.e.,

$$P(x_{j_1(i)}, x_{j_2(i)}, \cdots, x_{j_m(i)}) = \prod_{k=1}^{m} P(x_{j_k(i)}) \tag{5}$$

We seek to recover the structure of the GPT while minimizing (or completely eliminating) the need for external semantics to determine the directionality of the branches.

We first restrict the development to *non-degenerate* GPTs which, as we shall see in Section 3.4, does not limit practical application of the algorithm.

A probability distribution $P(x)$ is said to be non-degenerate if it has a *unique* skeleton, namely, every poly-tree representation of $P(x)$ must have the same set of branches (albeit different orientations). This implies that the GPT depicts each and every conditional independency embedded in $P$, i.e., whenever a set of instantiated nodes $S$ leaves an "unblocked" path between node $x_i$ and $x_j$, then $x_i$ and $x_j$ correspond to a pair of variables that are dependent (in $P$) given $S$ [2]. In DAGs, a path $p$ is said to be unblocked by a set $S$ of nodes if

1.  No arrow along $p$ emanates from a node in $S$, and

2.  Every node with converging arrows along $p$ is in $S$ or has a descendant in $S$.

In terms of the information measure $I$, non-degeneracy implies that for any pair of variables $(x_i, x_j)$ which do not have a common descendant we have

$$I(x_i, x_j) > 0 \tag{6}$$

and, similarly, if $x_k$ renders the path between $x_i$ and $x_j$ unblocked, then

$$I(x_i, x_j \mid x_k) > 0 \tag{7}$$

where

$$I(x_i, x_j \mid x_k) = \sum_{x_i, x_j, x_k} P(x_i, x_j, x_k) \log \frac{P(x_i, x_j \mid x_k)}{P(x_i \mid x_k) P(x_j \mid x_k)} \tag{8}$$

Note that, by definition (5), the set of parents of any variable are mutually indpendent, hence,

$$I(x_{j_1(i)}, x_{j_2(i)}) = 0 \tag{9}$$

The algorithm developed here will recover directionality to the maximum extent permitted by $P(X)$. That total directionality may not always be recoverable is apparent from examining the JPDFs of the three possible types of adjacent triplets allowed in poly-trees.

Type 1.        A -->-- B -->-- C

Type 2.        A --<-- B -->-- C

Type 3.        A -->-- B --<-- C

Since $P(A,B,C) = P(C|B)\,P(B|A)\,P(A) = P(C|B)\,P(A|B)\,P(B)$, the *JPDF*s of Types 1 and 2 are indistinguishable. For Type 3, however, we have $P(A,B,C) = P(B|A,C)\,P(A)\,P(C)$ which allows it to be uniquely identified since A and B are marginally independent and all other pairs are dependent. Given a skeleton tree, these relationships can be used to determine if a variable has multiple parents, and once the first two parents have been found, to then resolve the identity of all other parents and children. Specifically, we note that the partially directed triplet A -->-- B ----- C can be completed by testing for the mutual independence of A and C; if $A$ and $C$ are independent, $C$ is a parent of $B$, else, $C$ is a child of $B$.

These relationships permit the recovery of directionality in all the *causal basins* of a GPT. A causal basin starts with a multi-parent cluster (a child node and all of its direct parents) and continues in the direction of causal flow to include all of the child's descendants and all the direct parents of those descendants. Figure 1 shows an example of a poly-tree with two disjoint causal basins. As will become apparent, the directionality of a branch can be recovered if and only if it is contained within some causal basin of the generating poly-tree.

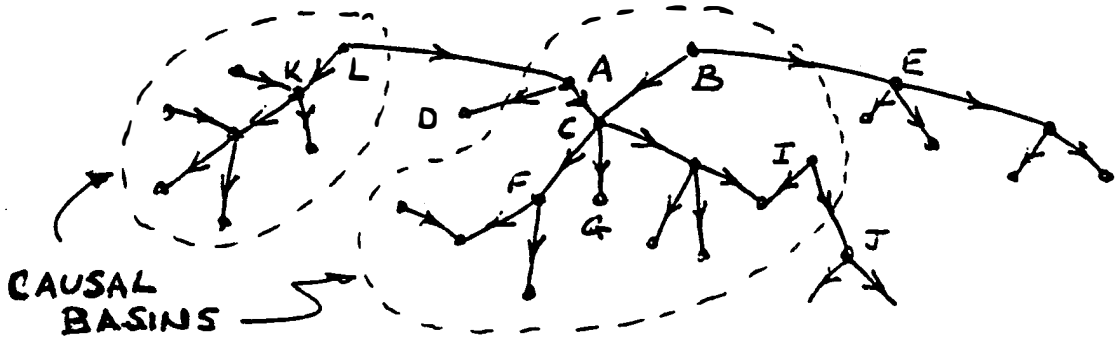

*Figure 1.*

## 3.2 Summary of the GPT Recovery Algorithm

The GPT recovery algorithm may be summarized in the following steps:

1.    Generate a skeleton (undirected spanning tree) by using the *MWST* procedure (steps 1 through 4) of Section 2.2.

2.    Search the internal nodes of the skeleton, beginning with the outermost layer and working inward, until a multi-parent node is found using the Type-3 test described in Section 3.1.

3.    When a multi-parent node $n$ is found, determine the directionality of all its branches using the Type-3 test.

4. For each node having at least one incoming arrow, use the partially-directed-triplet test to resolve the directionality of all its remaining adjacent branches.

5. Repeat steps 2 through 4 until no further directionality can be discovered.

6. If there remain undirected branches, label them "undetermined" and supply external semantics needed for completion.

7. From $P(x)$ compute the conditional probabilities prescribed in (4).

## 3.3 Theoretical Basis for GPT Recovery Algorithm

The theoretical basis for the GPT recovery algorithm stands on the following two theorems.

**Theorem 1.** The *MWST* algorithm of Section 2.2 unambiguously recovers the topology of any non-degenerate poly-tree.

**Proof:** Non-degeneracy implies, as in Eq.(6), that all branches of the GPT have non-zero weight and, in addition, that the conditional weights $I(A, C \mid B)$ are also non-zero for any pair $(A, C)$ not mediated by $B$. If $A$, $B$ and $C$ are any three variables obeying the conditional independence,

$$P(A \mid BC) = P(A \mid B),$$
(10)

then it is well known (Gallager [5]) that

$$I(A, B) = I(A, C) + I(A, B \mid C)$$
(11)

$$I(B, C) = I(A, C) + I(C, B \mid A)$$
(12)

Consequently, for any triplet $(A, B, C)$ satisfying 10, (e.g., Type-1 and Type-2 triplets) we have:

$$\min[I(A, B), I(B, C)] > I(A, C).$$
(13)

Equation (13) also holds for Type-3 triplets since $A$ and $C$ are marginally independent, thus, it holds for any triplet $A$, $B$, $C$ such that $B$ lies on the path connecting $A$ and $C$ in the GPT. Consequently, the *MWST* algorithm will never list a candidate branch weight of an unlinked pair $(x_i, x_j)$ higher than that of a legitimate GPT branch on the path connecting $x_i$ and $x_j$. Hence any attempt to select the unlinked pair $(x_i, x_j)$ would form a loop with already selected branches, and that would cause $(x_i, x_j)$ to be discarded. Therefore the skeleton recovered from $P(X)$ exactly matches that of the GPT.

**Theorem 2.** The directionality of a branch can be recovered if and only if it is contained within some causal basin of the poly-tree.

**Proof:** In the derived GPT skeleton each node with multiple neighbors is examined. The neighbors are pairwise tested to determine if at least two of them are marginally independent. Having found the first pair of parents lets us specify the remaining members of the multi-parent cluster. The (parent-descendant) identity of all members is the determined using the test for partially directed triplets (Section 3.1).

226

The succeeding generations of descendants are similarly resolved using the partially directed triplet test. This process is continued in the direction of causal flow, thus sweeping out the causal basin of the discovered multi-parent cluster. In a similar manner for any descendant we can identify all of its multiple parents which may first be encountered during the sweep of a given basin. However further 'upstream' causal resolution of such multiple parents is not possible unless they themselves have multiple parents.∥

**Corollary:** For simple trees (no multi-parent clusters) it is not possible to assign causal direction to any of the links without resorting to external semantics. This follows directly from the inability to identify any type-3 triplet and the indistinguishability of Type-1 from Type-2 triplets.

## 3.4 The Degenerate Case

Under conditions of degeneracy, $P(x)$ can be represented by several polytrees, each having a different skeleton. There is no way to guarantee then that the recovery algorithm will produce any particular polytree from this set. These conditions are normally reflected in equalities among branch weights, leading to ties in the construction of the maximum weight spanning tree. For example, if $P(x)$ restricts the variables $X$, $Y$ and $Z$ to be equal to each other, all branch weights are equal and $P$ can be represented by any of the following three skeleton trees

$$X—Y—Z \qquad Z—X—Y \qquad Y—Z—X.$$

The $MWST$ algorithm may produce any one of these skeletons, depending on the tie breaking rule used

We are still guaranteed, though, that at least one of the skeletons produced by the $MWST$ algorithm would permit a faithful representation of $P$ via the product expansion in (4). This is based on the fact that if $P$ can be represented by a set of $k$ distinct skeletons $T = (T_1, \cdots T_k)$ then each of these skeletons (and, perhaps others) must have maximum weight. The proof is similar to that of Theorem 1, except that (6), (7) and (13) may now permit equalities.

However, once a skeleton tree is found, the process of identifying and orientating its causal basins (steps 3 and 4 of Section 3.2) must now employ higher-order statistics. For example, parents can no longer be identified by merely having a zero $I$ measure because children, too, may be marginally independent. Hence, the criterion $I(A, C \mid B) > 0$ should be invoked to distinguish type-3 triplets from type-1 and type-2 triplets, for which the equality $I(A, C \mid B) = 0$ holds.

## 4. CONCLUSIONS

Poly-trees represent a much richer dependency models than trees, as their JPDFs are products of higher-order distributions. Yet, the proposed recovery method uses the efficient $MWST$ algorithm to recover skeletons of Poly-trees from second-order statistics. The directionality of the network is recovered to the maximum extent permitted by $P(X)$. The algorithm works best with GPTs rich in multi-parent clusters because these clusters provide the information needed for determining the directionality in their associated basins. Finally, a direct benefit of the algorithm is its ability to precisely pin-point the minimum (if any) external semantics required to determine the causal relationships among the variables considered.

## ACKNOWLEDGEMENTS

The authors would like to thank Norman Dalkey for reviewing the first draft of this manuscript and suggesting several improvements.

## REFERENCES

1. Pearl, J., "On Evidential Reasoning in a Hierarchy of Hypotheses," *Artificial Intelligence,* Vol. 28, No.1 (1986), pp.9-15.

2. Pearl, J., "Fusion, Propagation, and Structuring in Belief Networks", *Artificial Intelligence,* Vol. 29, (1986), pp.241-288.

3. Chow, C.K., and Liu, C.N., "Approximating Discrete Probability Distributions with Dependence Trees," *IEEE Transactions on Information Theory,* Vol. IT-14, No.3 (May 1968), pp.462-467.

4. Lewis, P.M., "Approximating Probability Distributions to Reduce Storage Requirements," *Information and Control,* (September 1959), vol 2, pp.214-225.

5. Gallager, R.G., "Information Theory and Reliable Communications", John Wiley and Sons, Inc., N.Y. 1968, Chapter 2, pp.13-27.